# Comparative analysis of evolutionary algorithms for image enhancement

## Anupriya Gogna

Email: anupriya13281@gmail.com

## Akash Tayal

ECE Department
Indira Gandhi Institute of Technology, GGSIP University,
Kashmere Gate, Delhi-110006, INDIA

**Abstract:** Evolutionary algorithms are metaheuristic techniques that derive inspiration from the natural process of evolution. They can efficiently solve (generate acceptable quality of solution in reasonable time) complex optimization (NP-Hard) problems. In this paper, automatic image enhancement is considered as an optimization problem and three evolutionary algorithms (Genetic Algorithm, Differential Evolution and Self Organizing Migration Algorithm) are employed to search for an optimum solution. They are used to find an optimum parameter set for an image enhancement transfer function. The aim is to maximize a fitness criterion which is a measure of image contrast and the visibility of details in the enhanced image. The enhancement results obtained using all three evolutionary algorithms are compared amongst themselves and also with the output of histogram equalization method.

## 1 Introduction

Enhancement of images to highlight details and improve contrast is a prerequisite in virtually all image processing applications. It is of importance in areas such as fault detection, computer vision, medical image analysis and remote sensing. The process of image enhancement involves transforming an input image to a form which provides better visual perception and is more suited for information retrieval.

Enhancement techniques can be divided into four categories based on the kind of transformation employed: point operations (e.g. histogram equalization), spatial operation (e.g. median filtering), transform operations (e.g. homomorphic filtering) and pseudocoloring (Jain, 1991). Traditional image enhancement techniques are empirical or heuristic in nature. They require human intervention for evaluating the quality of output image and tuning of algorithm parameters for improved results. However, automatic (unsupervised) image enhancement needs to proceed without such inputs. It requires defining an objective evaluation criterion to judge the suitability of enhanced image to the application under consideration. Also the algorithm must be capable of optimizing the parameters of transformation function based on the input image without human involvement. In the past several techniques have been proposed for automatic image enhancement (Hong, Wan and Jain, 1998; Shaked and Tastl, 2005; Tjahjadi and Celik, 2012).

Optimizing the algorithm and its parameters for every image type is an exhaustive task. Given the wide range of image types and their varied enhancement requirements, image enhancement becomes a complex optimization problem. It warrants the use of metaheuristic techniques (Talbi, 2009) which are capable of giving good results in a reasonable amount of time. In this paper we use evolutionary algorithms (Ashlock, 2006), a class of metaheuristic techniques, for automatic image enhancement. Evolutionary Algorithms (EA) are population based metaheuristic algorithms (Talbi, 2009) that are inspired by the biological process of evolution. They proceed by selecting individuals from the current population and combining them (mating) to produce offsprings as the generations proceed. Only the fittest individuals carry on to the next generation. EA include techniques such as Genetic Algorithm (GA) (Holland, 1975), Genetic Programming (GP) (Koza, 1992), Differential Evolution (DE) (Price, Storn and Lampinen, 2006) and Self Organizing Migration Algorithm (SOMA) (Zelinka, 2004). Several methods for gray level image enhancement and contrast improvement have been proposed based on evolutionary algorithms. Saitoh (1999) used genetic algorithm based approach for improving contrast of images. Pal (Pal, Bhandari and Kundu, 1994) employs GA to find an optimal set of parameters for the enhancement function. Differential evolution was used for adaptive image enhancement in (Yang, 2010). Munteanu and Lazarescu (1999) uses subjective evaluation criterion with genetic algorithm for the purpose of enhancement. Genetic algorithm used along with lifting wavelet scheme is employed for image enhancement in (Song et al., 2010).

In this paper we undertake comparative analysis of three evolutionary algorithms (namely GA, DE and SOMA) for the purpose of image enhancement. Comparison of EA based methods is also made with the technique of histogram equalization (HE). It is shown that SOMA performs better than the other techniques. Not only is the quality of enhanced image superior, the population size required for SOMA is also smaller than that required by other evolutionary algorithms. It also shows much less variance between different runs of the algorithm.

The rest of the paper is organized as follows. Section 2 gives details of the mathematical functions used for transformation and objective evaluation of transformed image. Description of evolutionary algorithms used in the paper is given in Section 3. In Section 4, the experimental setup and observations are highlighted. Conclusion and future work are discussed in Section 5.

## 2 Mathematical Function

Image enhancement involves transforming the input image so as to obtain results that have better quality than the unprocessed image. A high quality output image must possess a large dynamic range and a higher level of visible details. For automatic image enhancement, evaluation of enhanced image is done using an objective fitness criterion.

The purpose of optimization algorithm is to find the optimal set of parameters of transformation function which maximizes the fitness function.

*2.1 Transformation Function*

The method used in this paper for image enhancement belongs to the category of spatial operations (Gonzales and Woods, 1987). Spatial transformation operation can be represented as

$$v(i, j) = T[u(i,j)] \qquad (1)$$

where, $v(i, j)$ and $u(i, j)$ are the intensity values of $(i, j)^{th}$ pixel in the input and output image respectively. $T$ represents the transformation operator.

The transformation function (Munteanu and Rosa, 2004) used in this paper is an extension of statistical scaling technique (Jain, 1991). It is given in equation (2).

$$v(i,j) = \frac{k*M}{(\sigma(i,j)+b)}[u(i,j) - c \times \mu(i,j)] + \mu(i,j)^a \qquad (2)$$

In the above expression (equation 2),
$M$ (given by equation 3) is the average of intensity values of the entire input image.

$$M = \frac{1}{H \times V} \sum_{x=0}^{H-1} \sum_{y=0}^{V-1} u(x,y) \qquad (3)$$

$H \times V$ is the total number of pixels contained in the image (horizontal size × vertical size).

$\mu(i,j)$ (given by equation 4) is the local mean of the pixel values contained in the neighbourhood (window of size $n \times n$) of $(i, j)^{th}$ pixel

$$\mu(i,j) = \frac{1}{n \times n} \sum_{x=0}^{n-1} \sum_{y=0}^{n-1} u(x,y) \qquad (4)$$

$\sigma(i,j)$ (given by equation 5) is the local standard deviation of the pixel in a $n \times n$ neighbourhood of $(i, j)^{th}$ pixel

$$\sigma(i,j) = \left[\frac{1}{n \times n} \sum_{x=0}^{n-1} \sum_{y=0}^{n-1} [u(x,y) - \mu(i,j)]^2\right]^{0.5} \qquad (5)$$

and $a, b, c, k \in R^+$ are the parameters to be optimized. An optimal set of values of these four parameters is found using the optimization methodology used, for the image under consideration.

The transformation function given in equation (2) includes global as well as local information. Local level details and statistical information may become inconsequential in the global statistics of an image. Hence, a method using both global and local information, leads to a better enhancement of details than a method such as histogram equalization which uses just global information. In equation (2) last (additive) term has a smoothing or brightening effect on the image. A positive real value of constant b avoids division by zero in areas with zero standard deviation. A fraction of local mean value is subtracted from the pixel value based on constant c.

*2.2 Objective Evaluation Function*

The evaluation criteria for automatic image enhancement must be capable of quantifying image quality objectively. It should take into consideration all features of a (visually) good image such as high contrast, enhanced edges etc. We use a fitness criteria (given in equation 6) proposed in (Munteanu and Rosa, 2004).

$$F(sol) = \ln\left[ln[E(I_{s(sol)})]\right] \times \frac{n_{ep} \times I_{(sol)}}{H \times V} \times e^{H(I_{par})} \tag{6}$$

Where, *F(sol)* gives the fitness value for a parameter set *(a, b, c, and k)* value of *sol*. *I(sol)* is the enhanced image obtained using the transformation function (equation 2) with values of four parameters determined by *sol*. *Is(sol)* is the image obtained by applying sobel edge operator to the enhanced image, *I(sol)*. The expression for *Is* is given in equation 7.

$$I_s = [\delta h(i,j)^2 + \delta g(i,j)^2]^{0.5} \tag{7}$$

Where,

$$\delta g(i,j) = I(i-1,j+1) + 2I(i,j+1) + I(i+1,j+1) - I(i-1,j-1) - 2I(i,j-1) - I(i+1,j-1)$$

$$\delta h(i,j) = I(i+1,j+1) + 2I(i+1,j) + I(i+1,j-1) - I(i-1,j+1) - 2I(i-1,n) - I(i-1,j-1)$$

*E(Is)* is the sum of pixel values in *Is*

$n_{ep}$ is the number of edge pixels detected using sobel operator with automatic thresholding.

*H(I)* is a measure of entropy of image given by equation (8)

$$H = -\sum_{n=1}^{G} p_n \log(p_n) \tag{8}$$

*G* is the number of gray scale levels in the image, and $p_n$ is the frequency of pixel having the intensity *n*,

Above evaluation function ensures a high value of fitness function, for an image, if
- There are high number of edge pixels (i.e. higher number of edges are visible)
- There is greater intensity of pixels in edge image (i.e. edges are more visible and pronounced)
- There is high value of entropy (i.e. the information content of image is high and gray levels are evenly distributed, for more natural appearance)

Therefore, unlike histogram equalization, which concentrates on just distribution of gray scale levels, the methodology employed in this paper not only improves the contrast but also increase the visibility of the edges (makes edges sharper)

## 3 Evolutionary Algorithms

Evolutionary algorithms are population based metaheuristic algorithms. They start with an initial population of individuals (solutions in search space). Over the generations (iterations), individuals comprising the existing population are gradually replaced by new offsprings based on the chosen replacement strategy. Offsprings are produced by mating (crossover and mutation) of parents selected (based on their fitness score) from the current generation. EA rely on the concept of survival of the fittest for reaching the optimum. A description of evolutionary algorithms used in our analysis is given in this section.

### 3.1 Genetic Algorithm (GA)

Genetic algorithm was proposed by John Holland in 1975 (Holland, 1975). GA begins with an initial (randomly or heuristically generated) set of solutions (chromosome) spread over the search space (determined by problem under consideration). Based on the fitness function value of all individuals, some are selected to be parents. Commonly used selection mechanisms (Talbi, 2009) are elitism (fittest individuals selected), tournament selection (best individuals chosen from a randomly selected set), roulette wheel selection (selection probability proportional to fitness) and rank based criteria (relative fitness value assigned). Selected individuals combine via the process of crossover (mating) to produce offsprings. This causes offspring to inherit traits form both the parents. Some of the crossover operators (Ashlock, 2006) proposed in literature are uniform crossover, single point crossover (usually used with binary GA), and arithmetic crossover, geometric crossover (used with Real Coded GA). To introduce diversity in population, facilitating higher degree of explorative component, mutation is carried out. Mutation occurs with a low probability and brings about a few changes in some of the randomly selected individuals. From amongst the current population and offspring population, new generation is selected such that fittest individuals survive on to the next generation. The process proceeds till a suitable completion criterion (such as maximum number of generations or optimum cost) is met. The pseudocode for genetic algorithm is illustrated in figure 1.

**Figure 1**   Pseudocode for Genetic Algorithm

---

*Initialize population with randomly selected individuals (solution set)*
*Compute fitness of all chromosomes (individuals)*
***While termination criteria not met***
  1. *Select required number of parents based on population's fitness score*
  2. *Produce offspring using crossover operator*
  3. *Perform mutation selectively*
  4. *Evaluate fitness of offspring*
  5. *Select new generation*

***End While***

*3.2 Differential Evolution (DE)*

Differential evolution was proposed in 1995 by Storn and Price (Price, Storn and Lampinen, 2006). The methodology for DE is similar to that of GA with a few variations. It differs from GA in the use of recombination operator (rather than traditional crossover operators). Recombination operator produces offspring in the line joining the parents. It uses an arithmetic (linear combination) operation. Hence, DE is suited for individuals represented using real numbers rather than as binary string (as in traditional binary GA). The algorithm has two main tunable parameters: scaling factor F and crossover constant cr. Crossover constant controls the diversity in population while the scaling factor governs the amount of perturbation introduced. The convergence speed slows down for a high value of the scaling factor. The value of crossover constant has limited effect on the performance of algorithm. Next generation is selected from the parent and offspring population using fitness based selection mechanism. The current individual is replaced by the offspring if the latter has a higher fitness value. Several advancement and variations have been proposed in the past, which are presented in (Neri and Tirrone, 2010). The pseudocode for DE is given in figure 2.

**Figure 2**    Pseudocode for Differential Evolution

```
Initialize population with randomly generated individuals
While termination criteria not met
    For each individual of population, i
        Compute fitness of entire population f (pop_i)
        offspring ← Parent_1 + F × (Parent_2 − Parent_3)
        For each dimension (d) of individual (i)
            If rand < cr
                Off (i,d) ← offspring (i, d)
            else
                Off (i,d) ← parent (i, d)
            End if
        End for
        Evaluate fitness of offspring, f (off_i)
        If f (off_i) < population f (pop_i)
            pop_i ← off_i
        End if
    End For
End While
```

*3.3 Self-Organizing Migration Algorithm (SOMA)*

SOMA, introduced in 2000 by Zelinka (Zelinka, Lampinen and Nolle, 2001; Zelinka, 2004) is an evolutionary algorithm which derives inspiration from social behaviour of animals (e.g. grouping of animals while looking for food). Unlike other EAs it doesn't involve producing offspring at every generation. Instead a population of individuals move about in the search space. Each individual is characterized by its position which is updated in every migration loop. A migration loop of SOMA is equivalent to a generation of GA. Fitness of all individuals is computed and one with the highest fitness becomes the leader. All individuals move towards the leader in each migration loop. Apart from the dimensionality of problem and the population size, the parameters that govern this process are:

- Path Length: It decides the position at which an individual stops while following the leader. If path length is less than 1, individual while moving towards the leader stops short of the leader's position. If it is equal to 1, individual stops at the position of leader. For a path length greater than 1, the individual overshoots the leader's position and crosses over the latter (while moving in the direction of the leader). Path Length should be assigned a value greater than 1 to ensure extensive exploration of search space.
- Step Size: It decides the size of steps with which an individual traverse the path towards the leader, in one loop. A small value gives better results at cost of increased computation time.
- PRT: It is the pattern that governs the mutation in SOMA. A low value of PRT is usually recommended.

Based on the value of PRT, a *PRT_vector* (with dimension equal to the dimensionality of search space) is generated as shown in equation

$$PRT_{vector}(d) = \begin{cases} 1 & ; rand(d) < PRT \\ 0 & ; elsewhere \end{cases}; 1 \leq d \leq ND \qquad (9)$$

Where, *rand* is a vector (of size *ND*) of uniformly distributed random numbers between zero and one. This *PRT_vector* is responsible for inserting mutation component in the evolution process.

The position of each individual (except the leader) is updated in accordance with equation 10.

$$x_{n,d}^{ML+1} = x_{n,d,original}^{ML} + \left(x_{Leader,d}^{ML} - x_{n,d,original}^{ML}\right) \times t \times PRT\_vector_d \qquad (10)$$

Where, *t* is the current step, $= [0: step\_size: path\_length]$. $x_{n,d}^{ML+1}$ is the position of the *n*[th] individual (in the *d*[th] dimension) in the next migration loop, $x_{n,d,original}^{ML}$ is the original position of the *n*[th] individual (in the *d*[th] dimension) at the start of current migration loop. $x_{Leader,d}^{ML}$ is the position of the leader (in the *d*[th] dimension) in the current migration loop and *PRT_vector* is a ND-dimensional vector as given by equation 9.

**Figure 3**   Pseudocode for Self-Organizing Migration Algorithm

```
Generate a random initial population, pop
While termination criteria not met
    Compute fitness of each individual
    Select leader (individual with highest fitness)
    For each individual (i) of the population,
        While not path length
            Generate PRT_vector as illustrated in equation 9
            Update position of particle as given in equation 10
            Update fitness value
            Select a new leader
        End while
    End while
```

Several variants of SOMA have been proposed (Zelinka, 2004) apart from All to Leader (discussed above) such as All to rand (all individuals move to a randomly selected leader), All to All (individuals move to any randomly selected individual, no leader is there). The pseudocode for All to One (Leader) is given in figure 3.

## 4  Experiment and Results

Image enhancement was carried out on several images using the above mentioned evolutionary algorithms. Simulations of the proposed algorithms have been carried out using MATLAB 2010. The system configuration is Intel i5, 2.30 GHz processor with 3GB RAM. The images chosen have varied enhancement requirements. For example, "mandril" requires a greater focus on improvement of contrast, whereas the image "livingroom" needs enhancement of edges to make fine details better visible. "Cameraman" and "Lena" require both edge and contrast enhancement to be more pleasing visually. The boundary values for parameters a, b, c, and k are kept as (Munteanu and Rosa, 2004): $a \in [0, 1.5]$, $b \in [0, 1]$, $c \in [0, 0.5]$, $k \in [0.5, 1.5]$. Fitness value of enhanced images, visual inspection, convergence behavior of algorithms, variance between multiple runs of algorithm, and DV-BV (detail variance-background variance) of enhanced images are the parameters used for comparison of algorithms. The parameter values and kind of operators used for various algorithms are mentioned below. These parameters are chosen using trial and error procedures, no formal experimentation as carried out for the same.

### 4.1  Differential Evolution

Informal trial and error procedures were carried out to determine appropriate parameter sets for DE. Range of parameter values chosen for trials is: population size (30-100), mutation rate number of iterations (20-100), scaling factor (0.2-1.2) and crossover constant (0.1-0.8). Binomial crossover was used as it is the most widely used crossover mechanism for DE.

- Population Size: 60

- Dimension of each individual : 4
  Individuals are encoded as real valued vectors of length four wherein each value corresponds to one of the decision variables i.e *a, b, c, k.*
- Number of iterations: 50
- Crossover constant (cr): 0.2
- Scaling factor (F): It ranges from 0.4 to 1. Iterations start with a high value of F to facilitate exploration of large search space. As iterations proceeds, the value of F is reduced as per equation (11)

$$F(i) = [(1 - 0.4) \times (Maximum\ iterations - i)]/Maximum\ iterations \qquad (11)$$

- Crossover operator: binomial crossover (eq. 12)
  In case of binomial crossover each of the gene in the offspring is obtained from either the mutated vector (obtained from combination of 3 parents) or from the parent itself.

  *If rand(i) ≤ cr*
    *offspring(i) ← parent3(i) + F \*(parent1(i)− parent2(i))*
  *else*
    *offspring (i) ← parent(i)*
  *end if* (12)

### 4.2 Genetic Algorithm

Informal trial and error procedures were carried out to determine parameter sets for GA. Range of parameter values chosen for trials are: population size (30-100), mutation rate (0.01-0.1), number of iterations (20-100) and selection mechanism (tournament selection, elitism and roulette wheel selection). Arithmetic crossover was chosen as it is the most popular crossover strategy for real coded genetic algorithm.

- Population Size: 60
- Dimension of each individual : 4
  Individuals are encoded as real valued vectors of length four wherein each value corresponds to one of the decision variables i.e *a, b, c, k.* Thus we use real coded GA for our experiment.
- Number of iterations: 50
- Mutation rate: 0.03
- Crossover probability: 1.0
- Crossover: Arithmetic crossover (Michalewicz, 1996) given by equation (13)

$$offspring^1 = rand \times parent^1 + (1 - rand) \times parent^2$$
$$offspring^2 = rand \times parent^2 + (1 - rand) \times parent^1 \qquad (13)$$

- Selection Mechanism: Binary tournament selection with elitism.
  In this two individuals from the entire population are randomly chosen and the best amongst them is selected for mating. The process is repeated as many times as the desired number of parents. This scheme is used along with elitism where few best individuals from the population are selected for mating. In our case 6 best individuals are chosen by elitism and rest 54 are chosen by the method of binary tournament selection. Binary tournament selection exhibits high selection pressure. Use of K-Elitist scheme ensures that best individuals found during the search process are not lost over the generations.

*4.3 Self-Organizing Migration Algorithm*

Range and preferred parameter settings for SOMA was taken from (Zelinka, 2004)

- Population Size: 25
- Dimension of each individual : 4
  Individuals are encoded as real valued vectors of length four wherein each value corresponds to one of the decision variables i.e *a, b, c, k*.
- Number of iterations: 50
- PRT: 0.1
- Path length: 2
- Step Size: 0.21
- Mutation: Gaussian (Leandro, 2009) mutation indicated below. It gives better results than traditional mutation used with SOMA.

*If rand < 0.5*
$$\text{new}_{pos} \leftarrow \text{start}_{pos} + (\text{leader}_{pos} - \text{start}_{pos}) \times i \times \text{PRT}_{vector}$$
*else*
$$\text{new\_pos} \leftarrow \text{start\_pos} + \text{randn} \times (\text{leader\_pos} - \text{start\_pos}) \times i \times \text{PRT\_vector}$$
*End if*

(14)

*4.4 Results*

Various evolutionary algorithms and histogram equalization method have been compared using several criteria for the problem of image enhancement. These include fitness value (described in section 2.2), subjective visual inspection of enhanced images, distribution of gray scale levels, number of edge pixels and DV-BV values of the enhanced and original images. Algorithms are also compared in terms of their effectiveness, robustness and applicability to image enhancement problem. The results and conclusion are supported by statistical tests included in the analysis.

- *Fitness Value*

The evaluation function (fitness) value obtained for various algorithms is given in table 1 for four images. The values shown in the table are averages of fitness score of 35 runs of each algorithm.

**Table 1** Comparative fitness value for various algorithms
(Standard deviation is shown in brackets alongside)

|  | Original | HE | DE | GA | SOMA |
|---|---|---|---|---|---|
| Lena | 187.72 | 177.39 | 191.2 (0.31) | 268.46 (18.69) | 329.07 (0.98) |
| Cameraman | 127.13 | 83.67 | 171.31 (2.24) | 227.01 (2.65) | 262.08 (1.14) |
| Living Room | 162.89 | 137.38 | 228.8 (1.97) | 227.37 (10.94) | 241.88 (1.2) |
| Mandril | 105.24 | 101.32 | 162.98 (0.46) | 180.71 (13.38) | 205.81 (1.9) |

Fitness function values in table 1 indicate that HE (histogram equalization) results are poorer as it does not give importance to local information. Among the three evolutionary algorithms DE gives the lowest fitness score. GA in general performs better than DE in terms of average fitness score. SOMA is the best performing algorithm, giving the highest evaluation function value even with a much smaller population size (half of GA). Standard deviation of all algorithms is also mentioned. GA shows in general a lot of variability amongst runs. This arises due to strong dependence of performance of GA on the fitness of initial population. DE and SOMA on the other hand are relatively much more robust algorithms (with low standard deviation values).

- *Subjective Visual Inspection*

In figure 4 and 5 image "Mandril" and "Livingroom" are shown with their enhanced variants. In case of image "Mandril" the original image has a dull appearance. Enhancement of "Mandril" requires greater emphasis on improvement of contrast rather than on highlighting of details. This is because the image lacks too many details or features that needs to be emphasized upon.. Thus the focus shopuld be on improving gray level distribution while retaining the natural look of the image. The original image has a dull appearance, and the details are also not very sharp. In the image obtained using histogram equalization, although the contrast is enhanced, the visibility of details is not good. Also, the gray shades appear very dark. From amongst the evolutionary algorithms, DE gives results only slightly better than the original image. The enhancement of details (notice the lines near the nose and details of eyes), are most pronounced in SOMA.

**Figure 4**   Comparison of Enhanced Image "Mandril" using different methods

    4a. Original image      4b. Histogram Equalized image
    4c. DE enhanced image      4d. GA enhanced image
    4e. SOMA enhanced image

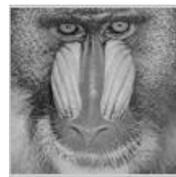 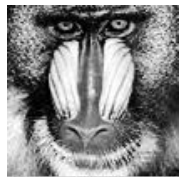 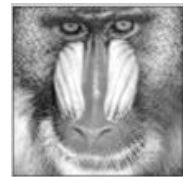

Fig. 4a. Original        Fig. 4b. HE        Fig. 4c. DE

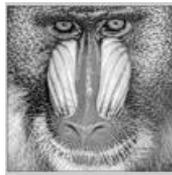 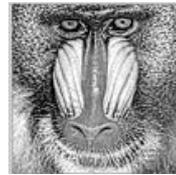

Fig. 4d. GA        Fig. 4e. SOMA

Another example is shown in figure 5 for the image "Livingroom". In contrast to the previous image, "Livingrom" has a large number of edges, features or fine details (edges and carving on door and table, frame on the window, etc.). Thus the focus of enhancement procedure in this case should be on highlighting the details rather than solely on improving image contrast. The result of histogram equalization has a deepened contrast, with intense shades of gray. However, the details are lost. On the other hand, results of metaheuristic techniques show a more soothing appearance and all the lines and edges are better visible as compared to the original image. Amongst the EAs, SOMA enhances the edges the most (notice the details of window pane)

Thus it can be seen observed from the visuals depicted in figure 4 and 5 that metaheuristic techniques are able to perform adequately and satisfactorily in situations requiring both contrast improvement and edge enhancement, two main requirements of image enhancement.

**Figure 5**    Comparison of Enhanced Image "Livingroom" using different methods

   5a. Original image        5b. Histogram equalized image
   5c. DE enhanced image    5d. GA enhanced image
   5e. SOMA enhanced image

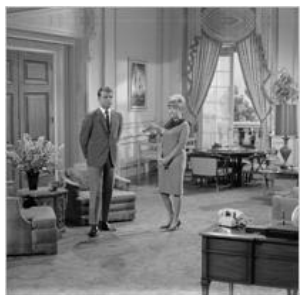 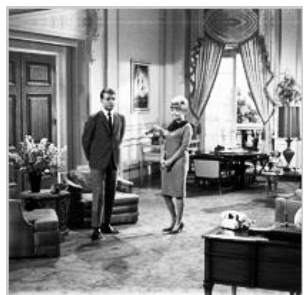 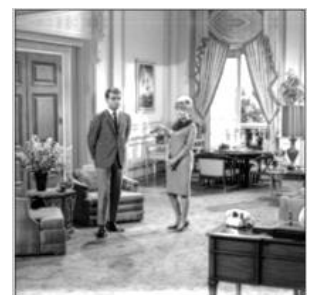

Fig. 5a. Original        Fig. 5b. HE        Fig. 5c. DE

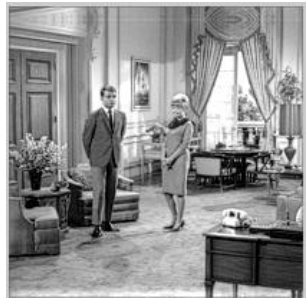 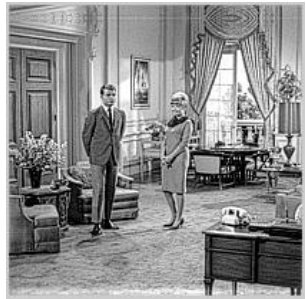

Fig. 5d. GA        Fig. 5e. SOMA

- *DV-BV Values*

Another criteria (Ramponi et al., 1996) used for the evaluation of images are the DV (Detail Variance) and BV (Background Variance) values. For computing DV and BV, firstly the variance of pixel values in the neighborhood (in our case, 3x3) of each pixel is computed. A threshold value (chosen to be 150) is selected, and all pixels with variance value greater than it are assigned to the foreground. The average of variance of all pixels in foreground give the DV. Variance values less than the thershold (background pixels) are averaged to give the BV. An increase in DV with relatively constant BV is an indication of improvement in image contrast. However, these paremeters are not the perfect evaluation criteria for judging the quality of enhanced image. The DV and BV values obtained for different images before and after enhancement are indicated in table 2. As can be seen, generally SOMA gives a much more significant improvement in DV (over the original image), as compared to other methodologies.

Table 2  Comparative Detail Variance (DV) and Background Variance (BV) for various algorithms

|           | Original | | HE | | DE | | GA | | SOMA | |
|-----------|------|------|------|------|------|------|------|------|------|------|
|           | DV   | BV   | DV   | BV   | DV   | BV   | DV   | BV   | DV   | BV   |
| Lena      | 938  | 14.7 | 1628 | 18.5 | 942  | 16   | 1133 | 18   | 1679 | 19   |
| Cameraman | 1600 | 7.4  | 1707 | 17.2 | 1738 | 7.7  | 2234 | 12   | 2220 | 17   |
| Livingroom| 612  | 17.4 | 1305 | 18.3 | 759  | 20   | 1073 | 23   | 1408 | 21   |
| Mandril   | 498  | 26.2 | 1414 | 28.2 | 589  | 28   | 1276 | 32   | 1553 | 33   |

- *Visiblity of Edges and Gray Level Distribution*

In figure 6, the edge image (detected using sobel operator with automatic thresholding) and histogram are shown for the image " livingroom" and its enhanced variants. An optimally enhanced image must posess a larger number of visible edges than the original and all the details must be clearly and prominently visible. Edge images are shown to showcase the extent to which each algorithm can enhance the edges. It can be observed from the given figure, SOMA enhanced image shows the highest number of edges. HE produces an edge image with detail visibility comparable to the original (not much improvement seen). To illustrate this edge visibilty a table of number of edge pixels in the enhanced image for all algorithms is also given in table 3. Another criteria used for evaluation is the image histogram. A high contrast image, with a uniform spread of gray scale (intensity) values has a flat (ideally uniform) histogram. As compared to the original image, histogram of all the enhanced images is much more flat (indicating more even distribution of intensity values). In this regard the method of HE surpasses evolutionary algorithms. From amongst all the evolutionary algorithms SOMA generates the flatest histogram. Thus, it can be stated that SOMA is the best performing algorithm in terms of achieving both contrast improvement and edge enhancement.

- *Convergence Behaviour of Algorithms*

The convergence behaviour and variability in diverse runs of the three evolutionary algorithm is shown in figure 7 (for the image "Lena"). Image "Lena" is chosen as it requires a balance between edge enhancement and contrast improvement for adequate improvement in the image. Thus, it is an appropriate candidate to represent and compare the behaviour of algorithms. Fig. 7 shows the graph of fitness function (evaluation criteria) value versus the number of iterations. Three curves (for three different runs) are shown. As is observed from the graphs below,

- GA suffers from premature convergence and the solution quality is a strong function of initial population and its fitness.
- As seen from fig. 7, GA shows a lot of variability amongst various runs, therefore it lacks robustness.
- DE does not generate good quality solutions as the final fitness value is not very high.
- The robustness of DE is better than that of GA. The variability amongst various curves is much lesser than in case of GA
- SOMA algorithms generate near optimal solutions (indicated by a high final fitness value).
- SOMA also displays good robustness (low variability amongst runs).

**Figure 6**    Edge Image and Histogram for image "Livingroom"

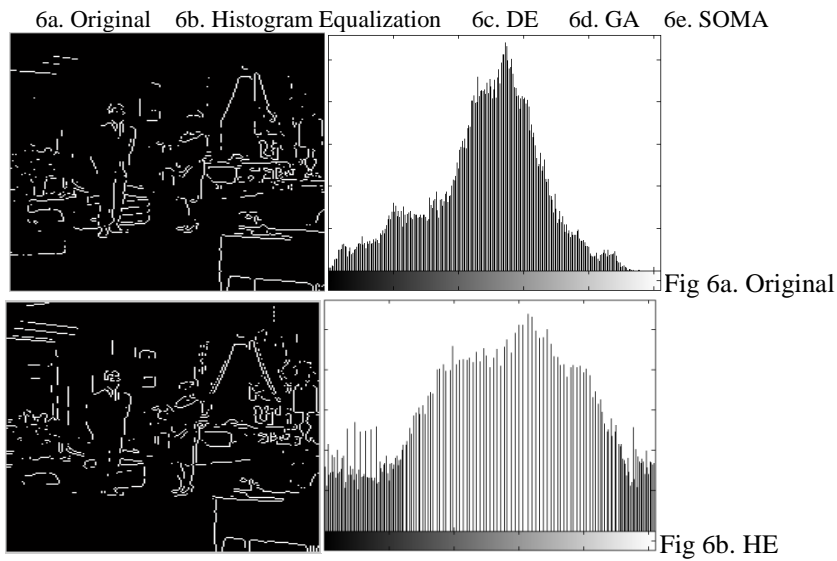

6a. Original    6b. Histogram Equalization    6c. DE    6d. GA    6e. SOMA

Fig 6a. Original

Fig 6b. HE

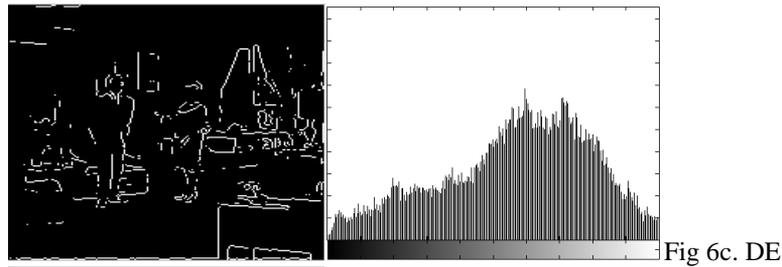
Fig 6c. DE

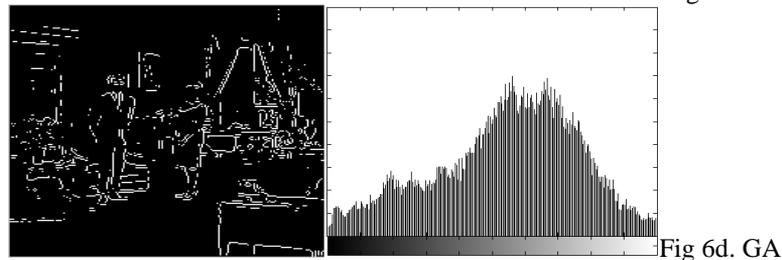
Fig 6d. GA

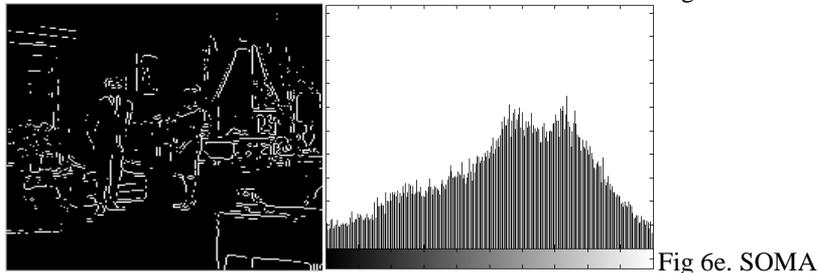
Fig 6e. SOMA

Therefore, SOMA is able to provide an optimum combination of good robustness and ability to reach optimum solution.

- *Simulation Time for Each Algorithm*

In this section simulation run times for each of the metaheuristic algorithm are enlisted. The values listed here are the average (standard deviation shown in brackets alongside) of 30 runs of each algorithm over all the problem instances.
- GA:     148.6 seconds (1.52)
- DE:     149.3 seconds (2.11)
- SOMA:   351.2 seconds (7.12)

It is seen that SOMA takes 2.5 times more time than other. However, SOMA is able to generate a fitness value much higher than that obtained using GA and DE as seen from table 1. Also, from the visual inspection of images, it can be seen that SOMA is able to more effectively improve contrast (figure 4) and enhance details (figure 5 and table 3) than other methods. Uniformity of gray scale distribution achieved using SOMA is also higher than that obtained using other metaheuristic techniques as indicated by histogram shown in figure 6. Also, figure 7 is an indication that robustness shown by SOMA is also fairly high; it shows very low variability amongst multiple runs. This is also supported by low values of standard deviation illustrated in table 1. Thus it can be stated that although SOMA requires greater simulation time than other evolutionary algorithms its performance in terms of image enhancement (as highlighted above) justifies its use. SOMA achieves much better performance than other algorithms at the cost of increased computation time. Use of dedicated processors can lead to reduction in algorithm's run times while still giving the benefits of SOMA aiding in real time processing.

- *Statistical Analysis of Results*

To prove that SOMA indeed performs better than the other two evolutionary algorithms, statistical tests were conducted on the fitness values obtained using all three evolutionary algorithms.

Firstly Kolomogorov-Smirnov test was done on the fitness score of each algorithm. The result indicated that all EAs yield results that follow a non-gaussian distributions. Hence, a non-parametric Kruskal-Wallis test was conducted to ascertain the difference between the fitness score of each of the evolutionary algortithms compared in this paper. The results (p-values) are indicated in table 4.

**Table 4** p-Values for pair of algorithms obtained using Kruskal-Wallis test

|            | DE-GA    | GA-SOMA  | DE-SOMA  |
|------------|----------|----------|----------|
| Lena       | 7.11E-05 | 1.6E-04  | 7.11E-05 |
| Cameraman  | 1.57E-04 | 1.57E-04 | 1.57E-04 |
| LivingRoom | 0.2568   | 0.0025   | 1.57e-04 |
| Mandril    | 7.11E-05 | 1.22E-04 | 7.11E-05 |

From the values indicated in table 4, it is evident that SOMA gives results that are statistically significantly different from those obtained using DE and GA (claim supported by a low associated p-value). Apart from the case of image "Livingroom", DE and GA also generate results that are significantly different.

**Figure 7** Convergence behaviour of evolutionary algorithms for multiple runs

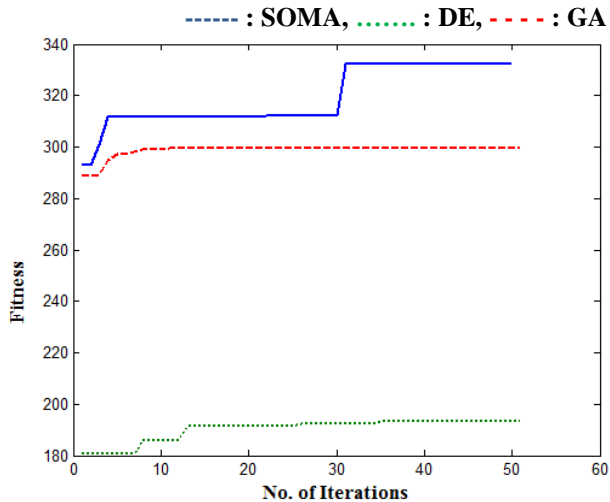

Fig. 7a. 1st run

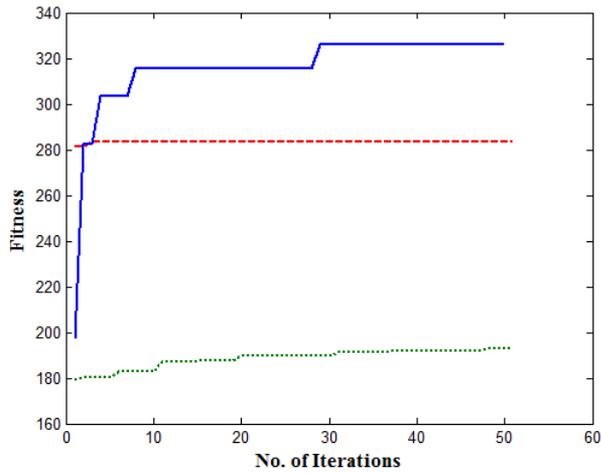

Fig. 7b. 2nd run

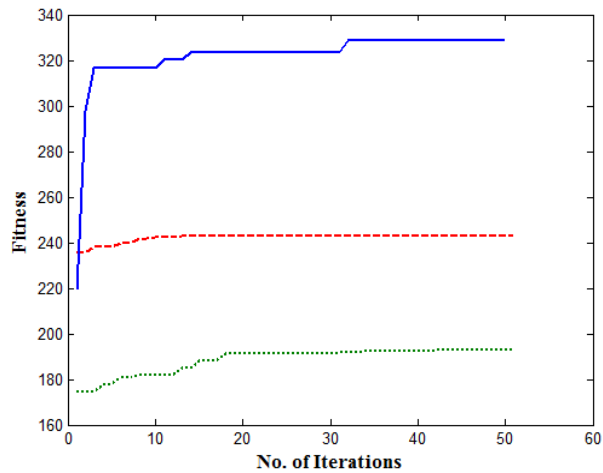

Fig. 7c. 3rd run

## 5 Conclusion and Future Work

In this paper the performance of three evolutionary algorithms (viz. Differential Evolution, Genetic Algorithm and Self Organizing Migration Algorithm) is compared for the problem of automatic gray scale image enhancement. The aim of optimization algorithm was to enhance the details while preserving the natural appearance of the image. The results of all the algorithms used are tabulated.

It is observed that SOMA performs better than both GA and DE. The results obtained using SOMA are much better (greater enhancement of details) than those obtained using other two algorithms. Also, SOMA is able to generate a flatter histogram implying even gray scale distribution than the other two evolutionary algorithms. Also the convergence behaviour of SOMA is better. SOMA shows the least variability for numerous runs of the algorithm, giving similar results every time the algorithm is run. The population size required for SOMA is also less than that needed for GA and DE. Although the computation time required for SOMA is higher, its use if justified by the performance features highlighted above. However, the performance of SOMA shows high dependence on the parameter selection, requiring extensive experimentation to find suitable set of values.

The proposed work can be extended by analysing the impact of variations in tuneable parameters on the results. Also, several modifications can be introduced into the existing algorithms framework to improve their behaviour and quality of solution. A hybrid approach can be adopted to improve performance of evolutionary algorithms by combining them with trajectory methods for a more effective exploitation of search space.

## References

Ashlock, D. (2006) Evolutionary Computation for Modelling and Optimization, Springer

Gonzales, R.C., Woods, R.E. (1987) Digital Image Processing. New York: Addison-Wesley

Holland, J. (1975) Adaptation in Natural and Artificial systems, University of Michigan Press, Ann Anbor

Hong, L., Wan, Y., Jain A., (1998) Fingerprint image enhancement: algorithm and performance evaluation, IEEE Transactions on Pattern Analysis and Machine Intelligence, Vol. 20(8), pp. 777-789

Jain, A.K., (1991) Fundamentals of Digital Image Processing, 1ed, Englewood Cliffs, NJ: Prentice-Hall

Koza, J.R., (1992) Genetic Programming, MIT Press, Cambridge, MA

Leandro dos S.C., (2009) Self-organizing migration algorithm applied to machining allocation of clutch assembly, Mathematics and Computers in Simulation, Vol. 80, pp. 427–435

Michalewicz, Z., (1996) Genetic Algorithms + Data Structures = Evolution Programs, 1ed, Springer Verlag.

Munteanu, C., Lazarescu, V., (1999) Evolutionary contrast stretching and detail enhancement of satellite images, Proceedings of International Mendel Conference, Czech Republic., pp. 94-99

Munteanu, C., Rosa, A., (2004) Gray-Scale Image Enhancement as an Automatic Process Driven by Evolution, IEEE Transaction on System, Man and Cybernetics-Part B: Cybernetics, Vol. 34, No.2, pp. 1292-1298

Neri, F., Tirrone, V., (2010) Recent advances in differential evolution: a survey and experimental analysis, Artificial Intelligence Review, pp. 61–106

Pal, S.K., Bhandari D. and Kundu, M.K.,(1994) Genetic algorithms for optimal image Enhancement, Pattern Recognition Letters, Vol. 15, pp. 261-271

Price, K.V., Storn, R.M. and Lampinen, J.A., (2006) Differential evolution: A Practical Approach to Global Optimization, Springer

Ramponi, G., Strobel, N.,Mitra, S.K. and Yu, T.H.,(1996) Nonlinear Unsharp Masking Methods for Image Contrast Enhancement, Journal of Electronic Imaging,Vol.5 No.3, pp. 353- 366.

Saitoh, F., (1999) Image contrast enhancement using Genetic Algorithm, Proceedings of IEEE SMC, Tokyo, Japan, pp. 899–904.

Shaked, D., Tastl, I., (2005) Sharpness measure: towards automatic image enhancement, Proceeding of IEEE International Conference on Image Processing, Vol. 1, pp. 937-940

Song, C., Gao, B., Kan, L.,Liang, H., (2010) Image Enhancement Based on Improved Genetic Algorithm and Lifting Wavelet Method, Proceeding of Chinese Control and Decision Conference, pp. 893-895

Talbi, E.-G., (2009) Metaheuristic: from design to implementation, 1ed, John Wiley & Sons

Tjahjadi, T., Celik, T., (2012) Automatic Image Equalization and Contrast Enhancement Using Gaussian Mixture Modeling', IEEE Transactions on Image Processing, Vol. 21(1), pp. 145-156

Yang, Q. (2010) 'An Adaptive Image Contrast Enhancement based on Differential Evolution, Proceeding of 3rd International Congress on Image and Signal Processing, Vol. 2, pp. 631-634

Zelinka, I., Lampinen, J., Nolle, L., (2001) On the theoretical proof of convergence for a class of SOMA search algorithms, Proceedings of the International MENDEL conference on soft computing, Czech Republic, pp. 103–107.

Zelinka, I., (2004) SOMA-Self Organizing Migrating Algorithm, in Onwubolu, G. et al (Eds.), New optimization techniques in engineering, Springer, pp. 167-215